# Black-box Generation of Adversarial Text Sequences to Evade Deep Learning Classifiers

Ji Gao, Jack Lanchantin, Mary Lou Soffa, Yanjun Qi
University of Virginia
http://trustworthymachinelearning.org/

## ABSTRACT

Although various techniques have been proposed to generate adversarial samples for white-box attacks on text, little attention has been paid to black-box attacks, which are more realistic scenarios. In this paper, we present a novel algorithm, DeepWordBug, to effectively generate small text perturbations in a black-box setting that forces a deep-learning classifier to misclassify a text input. We employ novel scoring strategies to identify the critical tokens that, if modified, cause the classifier to make an incorrect prediction. Simple character-level transformations are applied to the highest-ranked tokens in order to minimize the edit distance of the perturbation, yet change the original classification. We evaluated DeepWordBug on eight real-world text datasets, including text classification, sentiment analysis, and spam detection. We compare the result of DeepWordBug with two baselines: Random (Black-box) and Gradient (White-box). Our experimental results indicate that DeepWordBug reduces the prediction accuracy of current state-of-the-art deep-learning models, including a decrease of 68% on average for a Word-LSTM model and 48% on average for a Char-CNN model.

## 1 INTRODUCTION

Deep learning has achieved remarkable results in the field of natural language processing (NLP), including sentiment analysis, relation extraction, and machine translation [17, 29, 30]. However, recent studies have shown that adding small modifications to test inputs can fool state-of-the-art deep classifiers, resulting in incorrect classifications [7, 28]. This phenomenon was first formulated as adding very small and often imperceptible perturbations on images, which could fool deep classifiers on image classification tasks. It naturally raises concerns about the robustness of deep learning systems, considering that they have become core components of many security-sensitive applications such as text-based spam detection.

Formally, for a given classifier $F$ and test sample $\mathbf{x}$, recent literature defined such perturbations as vector $\Delta \mathbf{x}$ and the resulting sample $\mathbf{x}'$ as an *adversarial* sample[7]:

$$\mathbf{x}' = \mathbf{x} + \Delta \mathbf{x}, \ \|\Delta \mathbf{x}\|_p < \epsilon, \ \mathbf{x}' \in \mathbb{X} \tag{1}$$
$$F(\mathbf{x}) \neq F(\mathbf{x}') \ \text{or} \ F(\mathbf{x}') = t$$

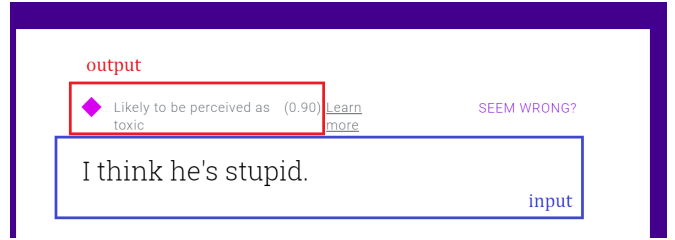

Figure 1: Perspective API: A example of deep learning text classification applications, which is a black-box scenario

Here we denote a machine learning classifier as $F : \mathbb{X} \rightarrow \mathbb{Y}$, where $\mathbb{X}$ is the sample space, $\mathbf{x} \in \mathbb{X}$ denotes a single sample and $\mathbb{Y}$ describes the set of output classes. The strength of the adversary, $\epsilon$, measures the permissible transformations. The choice of condition in Eq. (1) indicates two methods for finding adversarial examples: whether they are untargeted($F(\mathbf{x}) \neq F(\mathbf{x}')$) or targeted ($F(\mathbf{x}') = t$) [2].

The choice of $\Delta x$ is typically an $L_p$-norm distance metric. Recent studies [3, 7, 20, 28] used three norms $L_\infty$, $L_2$ and $L_0$. Formally for $\Delta \mathbf{x} = \mathbf{x}' - \mathbf{x} \in \mathbb{R}^d$, the $L_p$ norm is

$$\|\Delta \mathbf{x}\|_p = \sqrt[p]{\sum_{i=1}^{p} |x'_i - x_i|^p} \tag{2}$$

The $L_\infty$ norm measures the maximum change in any dimension. This means an $L_\infty$ adversary is limited by the maximum change it can make to each feature but can alter all the features by up to that maximum [7]. The $L_2$ norm corresponds to the Euclidean distance between $\mathbf{x}$ and $\mathbf{x}'$ [3]. This distance can still remain small when small changes are applied to many different features. An $L_0$ adversary is limited by the number of feature variables it can alter [20].

A third parameter for categorizing recent methods, in addition to targeted/untargeted and $\Delta$ choices, is whether the assumption of an adversary is black-box or white box. An adversary may have various degrees of knowledge about the model it tries to fool, ranging from no information to complete information. In the **black box** setting, an adversary is only allowed to query the target classifier and does not know the details of learned models or the feature representations of inputs. Since the adversary does not know the feature set, it can only manipulate input samples by testing and observing a classification model's outputs. In the **white box** setting, an adversary has access to the model, model parameters, and the feature set of inputs. Similar to the black-box setting, the adversary is still not allowed to modify the model itself or change the training data.

Figure 1 depicts the Perspective API[9] from Google, which is a deep learning based text classification system that predicts whether a message is toxic or not. This service can be accessed directly from the API website which makes querying the model uncomplicated and widely accessible. The setting is a black-box scenario as the model is run on cloud servers and its structure and parameters are not available. Many state-of-the-art deep learning applications have the similar system design: the learning model is deployed on the cloud servers, and users can access the model via an app through a terminal machine (frequently a mobile device). In such cases, a user could not examine or retrieve the inner structure of the models. Therefore, we believe that the black-box attack is generally more realistic than the white-box.

Most studies of adversarial examples in the literature use the white-box assumption [3, 7, 20, 28]. One study proposed by [19] showed that it is possible to create adversarial samples that successfully reduce the classification accuracy without knowing the model structure or parameters on image classification taks.

Recent studies have focused on image classification and typically created imperceptible modifications to pixel values through an optimization procedure [3, 7, 20, 28]. Szegedy et al. [28] first observed that DNN models are vulnerable to adversarial perturbation (by limiting the modification using $L_2$ norm) and used the Limited-memory Broyden-Fletcher-Goldfarb-Shanno (L-BFGS) algorithm to find adversarial examples. Their study also found that adversarial perturbations generated from one Convolutional Neural Network (CNN) model can also force other CNN models to produce incorrect outputs. Subsequent papers have explored other strategies to generate adversarial manipulations, including using the linear assumption behind a model [7] (by limits on $L_\infty$ norm), saliency maps [20] (by limits on $L_0$ norm), and evolutionary algorithms [18]. Recently, Carlini et al. proposed a group of attacking methods with optimization techniques to generate adversarial images with even smaller perturbations [3].

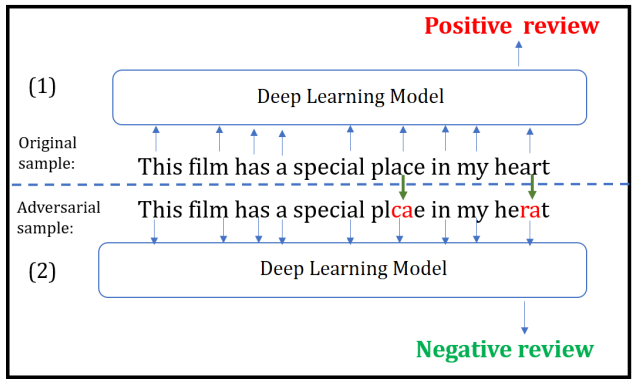

Figure 2: Example of a WordBug generated adversarial sequence. Part (1) shows an original text sample and part (2) shows an adversarial sequence generated from the original sample in Part (1). From Part (1) to Part (2), only a few characters are modified; however this fools the deep classifier to return a wrong classification.

In this study, we focus on generating adversarial samples on text data. Crafting adversarial examples on discrete inputs is fundamentally different from creating them for continuous inputs. Continuous input such as images can be naturally represented as points in a continuous $\mathbb{R}^d$ space ($d$ denotes the total number of pixels in an image). Using an $L_p$-norm based distance metric to limit the modification on images appears natural and intuitive. However, for text inputs it is difficult to search for small text modifications because of the following reasons:

(1) Text input **x** is symbolic. Perturbation on **x** is hard to define.
(2) No metric has been defined to measure text difference. $L_p$-norms makes sense on continuous pixel values, but not on texts since they are discrete.

Due to these reasons, the original definition of adversarial modifications (from Equation 1): $\Delta \mathbf{x} = \mathbf{x}' - \mathbf{x}$ cannot be applied directly to text inputs. Shown in Figure 3, one feasible definition of adversarial modifications on text can be the edit distance between text **x** and text **x**′ that is defined as the minimal edit operations that are required to change **x** to **x**′.

Multiple recent studies [21, 25] defined adversarial perturbations on RNN-based text classifiers. [21] first chose the word at a random position in a text input, then used a projected Fast Gradient Sign Method to perturb the word's embedding vector. The perturbed vector is projected to the nearest word vector in the word embedding space, resulting in an adversarial sequence (adversarial examples in the text case). This procedure may, however, replace words in an input sequence with totally irrelevant words since there is no hard guarantee that words close in the embedding space are semantically similar. [25] used the "saliency map" of input words and complicated linguistic strategies to generate adversarial sequences that are semantically meaningful to humans. However, this strategy is difficult to perform automatically.

We instead design scoring functions to adversarial sequences by making small edit operations to a text sequence such that a human would consider it similar to the original sequence. The small changes should produce adversarial words which are imperceptibly different to the original words. We do this by first targeting the important tokens in the sequence and then executing a modification on those tokens (defined in Section 2) that can effectively force a deep classifier to make a wrong decision. An example of adversarial sequence we define is shown in Figure 2. The original text input is correctly classified as positive sentiment by a deep RNN model. However, by changing only a few characters, the generated adversarial sequence can mislead the deep classifier to a wrong classification (negative sentiment in this case).

**Contributions:** This paper presents an effective algorithm, DeepWordBug (or WordBug in short), that can generate adversarial sequences for natural language inputs to evade deep-learning classifiers. Our novel algorithm has the following properties:

- Black-box: Previous methods require knowledge of the model structure and parameters of the word embedding layer, while our method can work in a black-box setting.
- Effective: Using several novel scoring functions on eight real-world text classification tasks, our WordBug can fool two different deep RNN models more successfully than the state-of-the-art baselines (Figure 6). Empirically, we also find that adversarial examples generated by our method to fool one deep model can evade similar models (Figure 7).

- **Simple:** WordBug uses simple character-level transformations to generate adversarial sequences, in contrast to previous works that use projected gradient or multiple linguistic-driven steps.
- **Small perturbations to human observers:** WordBug can generate adversarial sequences that look quite similar to seed sequences.

We believe that the techniques we present on text adversarial sequences can shed light in discovering the vulnerability of using DNN on other discrete inputs like malwares.

## 2 DEEPWORDBUG

For the rest of the paper, we denote samples in the form of pair $(\mathbf{x}, y)$, where $\mathbf{x} = x_1 x_2 x_3 ... x_n$ is an input text sequence including $n$ tokens (each token could be either a word or a character in different models) and $y$ set including $\{1, ..., K\}$ is a label of $K$ classes. A machine learning model is represented as $F : \mathbb{X} \to \mathbb{Y}$, a function mapping from the input set to the label set.

### 2.1 Background

*2.1.1 Recurrent Neural Networks.* Recurrent neural networks (RNN) [24] are a group of neural networks that include recurrent structures, artificial neuron structures with loops, to capture the sequential dependency among items of a sequence. RNNs have been widely used and have been proven to be effective on various NLP tasks including sentiment analysis[27], parsing[26] and translation[1]. Due to their recursive nature, RNNs can model inputs of variable length and can capture the complete set of dependencies among all items being modeled, such as all spatial positions in a text sample. To handle the "vanishing gradient" issue of training basic RNNs, Hochreiter *et al.* [8] proposed an RNN variant called the Long Short-term Memory (LSTM) network that achieves better performance comparing to vanilla RNNs on tasks with long-term dependencies.

*2.1.2 Convolutional Neural Network.* Convolutional neural networks (CNN) [10] are another group of neural networks that include convolutional layers, a network layer designed to capture the connectivity pattern in the input, to better extract features. A convolutional layer includes a set of small filters, which is a small array with rectangle shape. During calculation, every filter slides around the layer input, and at every place does a multiplication with the layer inputs and forms a rectangle shape output. The final output of a convolutional layer is the combination of all the filter outputs.

CNN was initially designed to work for image tasks; however, it has recently been shown to work well on language inputs as well [30] and even achieve state-of-the-art performance.

*2.1.3 Word Embedding Models.* Machine learning models, when processing text, must first discretize a text into tokens. These tokens can then be fed into the model. Words are often used as the smallest unit for input to a model. A word embedding is a mapping that projects every word in a dictionary to a unique vector in some vector space. Such mappings transform the discrete representations of words into features on a continuous space, which is more conducive to use with machine learning models.

| | $x_1$ | $x_2$ | $x_3$ | $x_4$ | $x_5$ | $x_6$ | $x_7$ | $x_8$ | $x_9$ |
|---|---|---|---|---|---|---|---|---|---|
| $x =$ | this | film | has | a | special | place | in | my | heart |
| $x' =$ | $x_1$ | $x_2$ | $x_3$ | $x_4$ | $x_5$ | $x_6'$ | $x_7$ | $x_8$ | $x_9'$ |
| | this | film | has | a | special | plcae | in | my | herat |

**Figure 3: An example of black box adversarial sample**

Word embeddings have shown success in many NLP tasks [4, 16, 22]. Such models hold a dictionary and build the word embedding based on the dictionary. To be able to work with words not in the dictionary, word embedding based models often add a special entry in the dictionary called "out of vocabulary," or "unknown". Those models work well if the word embedding is well constructed and covers most words that occur in the data.

*2.1.4 Character Level Models.* While word-level models using word embeddings are dominant in NLP tasks [1, 5], it is often easier, and sometimes more effective, to start from scratch and directly train a model with characters as inputs [30]. In a character-based model, every character is treated as one token and usually represented as a one-hot encoded vector. [30] showed that when trained on large datasets, deep networks can use character level inputs to achieve state-of-the-art results.

### 2.2 Method: Black-box Generation of Adversarial Sequences

This paper focuses on the black-box untargeted attacks. By the phrase "black-box attack,", we assume the attacker cannot access the structure, parameters or gradient of the target model. This is a realistic setting, because most modern machine learning classifiers are deployed as a service to receive users' inputs and provide corresponding outputs.

In the typical white-box setting of adversarial generation scenarios, gradients are used to guide the modification of input tokens from an original sample to an adversarial sample. However, gradients are hard to define on symbolic text inputs. Also in black-box settings, calculating gradients is not possible since the model parameters are not observable.

Therefore, we design a method to generate adversarial modifications on the input tokens directly, without the guidance of gradients. Considering the vast search space of possible changes (among all words/characters changes in an input example), we propose a two step approach to crafting adversarial samples in the black-box setting:

- **Step 1:** Determine the important tokens to change.
- **Step 2:** Modify them slightly, creating "imperceivable" changes which can evade a target deep learning classifier.

Specifically, to find important tokens, we design scoring functions to evaluate which tokens are important for the target model to make its decision. These scoring functions are used to determine the importance of any word to the final prediction. Once the primary tokens have been ascertained, we design and use a simple but powerful algorithm to transform those tokens and form an adversarial sample.

## 2.3 Step 1: Token Scoring Function and Ranking

First, we construct scoring functions to determine which tokens are important for the prediction. Our key idea is to score the importance of tokens in a input sequence according to the classification from the classifier. The proposed scoring functions have the following properties:

(1) Able to correctly reflect the importance of words for the prediction.
(2) Work without relying the knowledge of the parameters and/or structure of the classification model.
(3) Efficient to calculate.

In the following, we explain the four scoring functions we propose: Replace-1 Score, Temporal Head Score, Temporal Tail Score, and Combination Score. We assume the input sequence $\mathbf{x} = x_1 x_2 ... x_n$, where $x_i$ represents the token at the $i^{th}$ position. In order to rank tokens by importance, we need to measure the effect of the $i^{th}$ token on the output classification.

### 2.3.1 Replace-1 Score (R1S).
In the continuous case (e.g., image), suppose a small perturbation changes $x_i$ to $x'_i$. The resulting change of prediction output $\Delta F(\mathbf{x})$ can be approximated using the partial derivative of this $i^{th}$ feature:

$$\Delta F(\mathbf{x}) = F(\mathbf{x}') - F(\mathbf{x}) \approx (x'_i - x_i) \frac{\partial F(\mathbf{x})}{\partial x_i} \quad (3)$$

However, in a black-box setting, $\frac{\partial F(\mathbf{x})}{\partial x_i}$ is not available. Therefore, we directly measure the effect of replacing token $x_i$ with $x'_i$ as the Replace-1 Score (R1S):

$$\text{R1S}(x_i) = F(x_1, x_2, ..., x_{i-1}, x_i, ..., x_n)$$
$$- F(x_1, x_2, ..., x_{i-1}, x'_i, ..., x_n)$$

However, the search space of possible $x'_i$ is still very large. Therefore, we choose $x'_i$ to be the token "*unknown*", or so-called "out of vocabulary." By measuring the effect of changing $x_i$ to be "*unknown*," we achieve a fair comparison of the importance of the tokens.

### 2.3.2 Temporal Head Score (THS).
Since RNNs model inputs tokens in a sequential (temporal) manner, it is logical to compute token importance by the sequential prediction. Therefore we define a temporal head score (THS) of the $i^{th}$ token in an input sequence $\mathbf{x}$ as the difference between the model's prediction score as it reads up to the $i^{th}$ token, and the model's prediction score as it reads up to token $i - 1$:

$$\text{TS}(x_i) = F(x_1, x_2, ..., x_{i-1}, x_i) - F(x_1, x_2, ..., x_{i-1})$$

The Temporal Head Score of every token in the input sequence can be calculated efficiently by a single forward pass in the RNN models.

### 2.3.3 Temporal Tail Score (TTS).
The problem with the Temporal Head Score is that it scores a token based on its preceding tokens. However, tokens following that specific token are also important for the purpose of classification. Therefore we define the Temporal Tail Score as the complement of the Temporal Head Score. It compares the difference between two trailing parts of a sentence, the one containing a certain token versus the one that does not. The difference reflects whether the token influences the final prediction

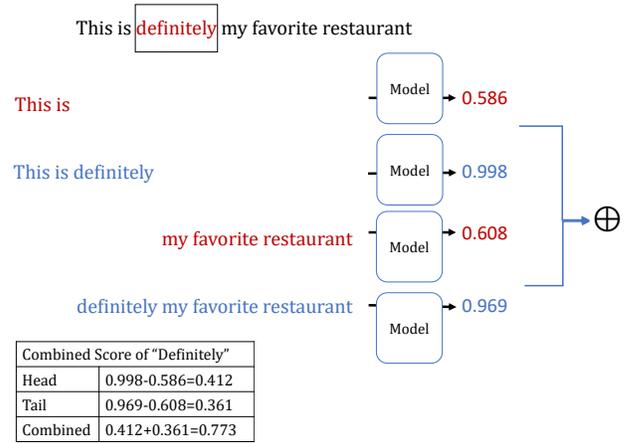

Figure 4: Combined score of "Definitely": A combination of Temporal Head Score and Temporal Tail Score

when coupled with tokens that followed. The Temporal Tail Score (TTS) of the token $x_i$ is calculated by:

$$\text{TTS}(x_i) = F(x_i, x_{i+1}, x_{i+2}, ..., x_n) - F(x_{i+1}, x_{i+2}, ..., x_n)$$

### 2.3.4 Combined Score (CS).
Since the Temporal Head Score and Temporal Tail Score both model the importance of a token from opposing directions, we can combine them to ascertain the importance of a token via its entire surrounding context. We calculate the Combined Score (CS) function as:

$$\text{CS}(x_i) = \text{THS}(x_i) + \lambda(\text{TTS}(x_i))$$

where $\lambda$ is a hyperparameter. The calculation of combined score is sketched in Figure 4.

Once we estimate the importance of each token in the input sequence, we select the top $m$ tokens to perturb in order to create an adversarial sequence.

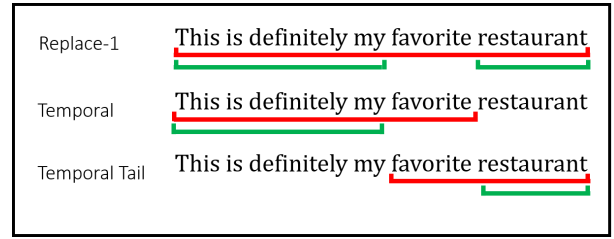

Figure 5: Illustration of scoring the token "favorite" in the input sequence "This is definitely my favorite restaurant". For each scoring method (Replace-1, Temporal Head, and Temporal Tail), the score of the token is equal to the prediction score of the red part minus the prediction score of the green part.

## 2.4 Step 2: Token Transformer

Any adversarial text generator needs a token transformer; that is, a mechanic to perturb text. Given the top $m$ important tokens from a scoring function, the second part of creating the adversarial sequence is to modify or perturb the tokens. For the character based model, the modification of a token can be done directly. That is, we can either substitute the token with a random one, or substitute it with a certain symbol such as space. However, for the word based

| Original | | Swap | Substitution | Deletion | Insertion |
|---|---|---|---|---|---|
| Team | → | Taem | Texm | Tem | Tezam |
| Artist | → | Artsit | Arxist | Artst | Articst |
| Computer | → | Comptuer | Computnr | Compter | Comnputer |

Table 1: Different transformer functions and their results.

models the transformation is more complicated, since the search space is large.

Previous approaches (summarized in Table 6) modify words following the gradient direction (gradient of the target adversarial class w.r.t the tokens) or following some perturbation guided by the gradient. However, there are no gradients available in a black-box scenario. Therefore, we propose efficient methods to modify a given word, and we do this by deliberately creating misspelled words. The motivation is similar to the image scenario where we want to make imperceptibly small changes to the input such that the the output classification changes. We define these small changes in text inputs as modifications of few individual characters.

The key observations are words are symbolic and learned classification models handle words through a dictionary to represent a finite set of possible words. The size of the typical NLP dictionary is much smaller than the possible combinations of characters. This means if we deliberately create misspelled words on important words, we can easily convert those important words to "*unknown*" (i.e., words not in the dictionary). Unknown words are mapped to the "*unknown*" embedding vector, which is likely vastly different than the embedding for the original word. Our results (Section 3) strongly indicate that this simple strategy can effectively force deep learning models to make wrong classifications.

Many strategies can be employed to create such misspellings. However, following the original definition of adversarial samples from [7], we prefer small changes to the original word as we want the generated adversarial sequence and its original sequence to appear (visually or morphologically) similar to human observers. Therefore, we use the Levenshtein distance (edit distance) [11], which is a metric that measures the similarity between sequences. While restricting the edit distance does not guarantee a cogent or coherent sentence meaning in a technical grammatic sense, it connects to the fact that humans have been shown capable of reading and comprehending sentences including a small number of typos, which has been researched in several psychological studies[23]. Setting restrictions on the maximum edit distance difference allows us to generate a sample that is readily understandable to human observers while tolerating a defined and limited number of typos.

We propose four similar methods:

(1) **Swap**: Swap two adjacent letters in the word.
(2) **Substitution**: Substitute a letter in the word with a random letter.
(3) **Deletion**: Delete a random letter from the word.
(4) **Insertion**: Insert a random letter in the word.

These transforms are shown in Table 1. The edit distance for the substitution, deletion and insertion operations is 1 and for the swap operation is 2.

These methods do not guarantee the original word is changed to a misspelled word. It is possible for a word to "collide" with another word after the transformation. However, the probability of collision is very small as there are $26^7 \approx 8 \times 10^9$ combinations for 7 letter words without hyphens and apostrophes, but a dictionary often includes no more than 50, 000 words, making the space very sparse.

| | #Training | #Testing | #Classes | Task |
|---|---|---|---|---|
| AG's News | 120,000 | 7,600 | 4 | News Categorization |
| Amazon Review Full | 3,000,000 | 650,000 | 5 | Sentiment Analysis |
| Amazon Review Polarity | 3,600,000 | 400,000 | 2 | Sentiment Analysis |
| DBPedia | 560,000 | 70,000 | 14 | Ontology Classification |
| Yahoo! Answers | 1,400,000 | 60,000 | 10 | Topic Classification |
| Yelp Review Full | 650,000 | 50,000 | 5 | Sentiment Analysis |
| Yelp Review Polarity | 560,000 | 38,000 | 2 | Sentiment Analysis |
| Enron Spam Email | 26,972 | 6,744 | 2 | Spam E-mail Detection |

Table 2: Dataset details

Different transformation algorithms can also be applied to the character-based models. In this case, the operation of modifying any token will lead to the difference of 1 on the edit distance.

The adversarial sample generation of DeepWordBug is summarized in Algorithm 1. From the algorithm, we could tell that generating adversarial samples using DeepWordBug takes at most $O(n)$ queries to the original model, where $n$ is the sequence length.

---
**Algorithm 1** DeepWordBug Algorithm

**Input:** Input sequence $\mathbf{x} = x_1 x_2 \ldots x_n$, RNN classifier $F(\cdot)$, Scoring Function $S(\cdot)$, Transforming function $T(\cdot)$, maximum allowed perturbation on edit distance $\epsilon$.

1: **for** $i = 1..n$ **do**
2:   $scores[i] = S(x_i; \mathbf{x})$
3: **end for**
4: Sort $scores$ into an ordered index list: $L_1 .. L_n$ by descending score
5: $\mathbf{x}' = \mathbf{x}$
6: $cost = 0, j = 1$
7: **while** $cost < \epsilon$ **do**
8:   $cost = cost + \text{Transform}(x'_{L_j})$
9:   $j + +$
10: **end while**
11: Return $\mathbf{x}'$
---

## 3 EXPERIMENTS ON EFFECTIVENESS OF ADVERSARIAL SEQUENCES

We evaluated the effectiveness of our algorithm by conducting experiments on different deep learning models across several real-world NLP datasets. In particular, we wanted to answer the following research questions:

- Does the accuracy of deep learning models decrease when fed the adversarial samples?
- Do the adversarial samples generated by our method transfer between models?
- Are DeepWordBug strategies robust to configuration parameters, such as dictionary size or transformer choices?

|  | Character Inputs | | Word Inputs | |
| --- | --- | --- | --- | --- |
| Dataset | Mean | Median | Mean | Median |
| AG's News | 602.96 | 596 | 40.6 | 40 |
| Amazon Review Full | 657.87 | 626 | 82.61 | 75 |
| Amazon Review Polarity | 657.45 | 628 | 82.57 | 74 |
| DBPedia | 639.39 | 623 | 54.48 | 56 |
| Yahoo! Answers | 623.44 | 623 | 85.2 | 57 |
| Yelp Review Full | 694.64 | 649 | 130.8 | 98 |
| Yelp Review Polarity | 697.03 | 649 | 129.53 | 101 |
| Enron Spam | 772.91 | 748 | 190.03 | 150 |

Table 3: Dataset Sample Length Statistics. Mean and median words (or characters) per sample in each dataset.

## 3.1 Experimental Setup

We performed 4 experiments to answer our research questions. Our experimental set up follows:

- **Datasets:** In our experiments, we use 7 large-scale text datasets from [30] together with the Enron Spam Dataset [15], which includes a variety of NLP tasks, e.g., text classification, sentiment analysis and spam detection. Details of the datasets are listed in Table 2.
- **Target Models:** To show that our method is effective, we performed our experiments on two well trained models: 1. Word-level LSTM (Word-LSTM), and 2. Character-level CNN (Char-CNN).

  **Word-LSTM:** The Word-level LSTM is a Bi-directional LSTM, which contains an LSTM in both directions (reading from first word to last and from last word to first). The network contains a random embedding layer to accept the word inputs. The embedding vectors are then fed through two LSTM layers (one for each direction) with 100 hidden nodes each. The hidden states of each direction are concatenated at the output and fed to a fully connected layer for the classification.

  **Char-CNN:** We use the same character-level CNN from [30] which uses one-hot encoded characters as inputs to a 9-layer convolutional network.

  The performance of the models without adversarial samples is presented in Table 4. In the non-adversarial setting, these models are similar to state-of-the-art results on these datasets.
- **Platform:** We train the target deep-learning models and implement attacking methods using software platform PyTorch 0.3.0. All the experiments run on a server machine, whose operating system is Ubuntu 14.04 and have 4 Titan X GPU cards.
- **Evaluation:** Performance of the attacking methods is measured by the accuracy of the deep-learning models on the generated adversarial sequences. The lower the accuracy, the more effective is the attacking method. Essentially it indicates the adversarial samples can successfully fool the deep-learning classifier model. The maximum allowed perturbation $\epsilon$ is a hyperparameter, measured using edit distance.

## 3.2 Methods in comparison

To evaluate our white-box scoring functions, we implemented two other methods as baselines and compared them to our proposed method. All three methods use the same transformer function because we assume that the selected words should be modified to minimize edit distance. In total, the 3 methods are:

(1) **Random (baseline):** This scoring function randomly selects tokens as targets. In other words, it has no method to determine which tokens to attack.
(2) **Gradient (baseline):** Contrary to random selection which uses no knowledge of the model, we also compare to full knowledge of the model, where gradients are used to find the most important tokens. Following equation 3, the gradient method uses the size of gradient w.r.t the original classification to decide which tokens should be changed. This method of selecting tokens is proposed in [25].
(3) **DeepWordBug (our method):** We use our white-box scoring functions to find the most important tokens. In our implementations, we use our 4 different scoring functions: Replace-1 Score, Temporal Head Score, Temporal Tail Score and the Combined Score.

| Dataset \Model | Char-CNN | LSTM | BiLSTM |
| --- | --- | --- | --- |
| AG's News | 89.96 | 88.45 | 90.49 |
| Amazon Review Full | 61.10 | 61.96 | 61.97 |
| Amazon Review Polarity | 95.20 | 95.43 | 95.46 |
| DBPedia | 98.37 | 98.65 | 98.65 |
| Yahoo! Answers | 71.01 | 73.42 | 73.40 |
| Yelp Review Full | 63.46 | 64.86 | 64.69 |
| Yelp Review Polarity | 95.26 | 95.87 | 95.92 |
| Enron Spam | 95.63 | 96.92 | 96.4 |

Table 4: Models' accuracy in the non-adversarial setting

For the Word-LSTM model, we use our 4 different transformers to change the words: substitution, deletion, insertion, swap. For the character model, we use the substitution transformer to change the words.

## 3.3 Experimental Results on Classification

We analyze the effectiveness of the attacks on the two models (Word-LSTM, Char-CNN) across eight datasets.
**Main Results:**

Model accuracy results on all 8 datasets when the maximum edit distance difference is limited to 30 ($\epsilon = 30$) are summarized in Table 5. It is clear that when modifying at most 30 characters in the input sequence, our method successfully generates samples that cause state-of-the-art deep learning classifiers to lose much accuracy, thus successfully evading the classifier. Figure 6 (b)(d) shows a direct comparison of the effectiveness of our attack to the baseline methods: Suppose the original classification accuracy as standard of performance, DeepWordBug reduce 68% performance of the Word-LSTM model and 48% performance of the Char-CNN model, which is much better than baselines in comparison.

For the Word-LSTM model, when $\epsilon = 30$ DeepWordBug with Combined Scoring and Substitution Transformer reduce model accuracy from 90% to around 25% on the AG's News Dataset and from 95% to around 36% on the Amazon Review Polarity Dataset. For the Char-CNN model, when $\epsilon = 30$ DeepWordBug with Replace-1 scoring and Substitution Transformer can reduce the model accuracy from 90% to around 30% on the AG's News Dataset and from 95% to 46% on the Amazon Review Polarity Dataset.

From Table 3, we can see that the mean and median sample lengths for each dataset are much larger than the number of words and characters modified. For example, for the Word-LSTM model on the Enron Spam Dataset, we changed only 16% of the words

## Word-LSTM Model

| | Baselines | | | | WordBug | | | | | | | |
|---|---|---|---|---|---|---|---|---|---|---|---|---|
| | Original | Random | | Gradient | | Replace-1 | | Temporal Head | | Temporal Tail | | Combined | |
| | Acc(%) | Acc(%) | Decrease | Acc(%) | Decrease | Acc(%) | Decrease | Acc(%) | Decrease | Acc(%) | Decrease | Acc(%) | Decrease |
| AG's News | 90.5 | 89.3 | 1.33% | 48.5 | 10.13% | 36.1 | 60.08% | 42.5 | 53.01% | 21.3 | 76.48% | 24.8 | 72.62% |
| Amazon Review Full | 62.0 | 61.1 | 1.48% | 55.7 | 10.13% | 18.6 | 70.05% | 27.1 | 56.30% | 17.0 | 72.50% | 16.3 | 73.76% |
| Amazon Review Polarity | 95.5 | 93.9 | 1.59% | 86.9 | 8.93% | 40.7 | 57.36% | 58.5 | 38.74% | 42.6 | 55.37% | 36.2 | 62.08% |
| DBPedia | 98.7 | 95.2 | 3.54% | 74.4 | 24.61% | 28.8 | 70.82% | 56.4 | 42.87% | 28.5 | 71.08% | 25.3 | 74.32% |
| Yahoo! Answers | 73.4 | 65.7 | 10.54% | 50.0 | 31.83% | 27.9 | 61.93% | 34.9 | 52.45% | 26.5 | 63.86% | 23.5 | 68.02% |
| Yelp Review Full | 64.7 | 60.9 | 5.86% | 53.2 | 17.76% | 23.4 | 63.83% | 36.6 | 43.43% | 20.8 | 67.85% | 24.4 | 62.28% |
| Yelp Review Polarity | 95.9 | 95.4 | 0.55% | 88.4 | 7.85% | 37.8 | 60.63% | 70.2 | 26.77% | 34.5 | 64.04% | 46.2 | 51.87% |
| Enron Spam Email | 96.4 | 67.8 | 29.69% | 76.7 | 20.47% | 39.1 | 59.48% | 56.3 | 41.61% | 25.8 | 73.22% | 48.1 | 50.06% |
| Mean | | | 6.82% | | 16.46% | | **63.02%** | | 44.40% | | **68.05%** | | **64.38%** |
| Median | | | 2.57% | | 13.95% | | **61.28%** | | 43.17% | | **69.46%** | | **65.15%** |
| Standard Deviation | | | 9.81% | | 8.71% | | 4.94% | | 9.52% | | 6.77% | | 9.56% |

## Char-CNN Model

| | Baselines | | | | WordBug | | | | | | | |
|---|---|---|---|---|---|---|---|---|---|---|---|---|
| | Original | Random | | Gradient | | Replace-1 | | Temporal Head | | Temporal Tail | | Combined | |
| | Acc(%) | Acc(%) | % Decrease | Acc(%) | Decrease | Acc(%) | Decrease | Acc(%) | Decrease | Acc(%) | Decrease | Acc(%) | Decrease |
| AG's News | 90.0 | 82.4 | 8.36% | 62.3 | 30.74% | 30.8 | 65.80% | 74.1 | 17.66% | 58.6 | 34.90% | 60.4 | 32.88% |
| Amazon Review Full | 61.1 | 51.0 | 16.53% | 47.0 | 23.04% | 25.6 | 58.17% | 58.1 | 4.89% | 32.5 | 46.79% | 35.0 | 42.70% |
| Amazon Review Polarity | 95.2 | 93.4 | 1.91% | 84.3 | 11.41% | 46.4 | 51.27% | 91.6 | 3.79% | 70.9 | 25.48% | 73.5 | 22.83% |
| DBPedia | 98.4 | 95.8 | 2.58% | 92.9 | 5.60% | 74.9 | 23.91% | 95.7 | 2.73% | 88.2 | 10.37% | 88.8 | 9.69% |
| Yahoo! Answers | 71.0 | 52.2 | 26.45% | 43.5 | 38.76% | 30.0 | 57.72% | 56.8 | 20.05% | 35.3 | 50.23% | 36.6 | 48.50% |
| Yelp Review Full | 63.5 | 52.6 | 17.05% | 45.7 | 28.06% | 27.6 | 56.56% | 51.3 | 19.10% | 35.3 | 44.36% | 38.2 | 39.74% |
| Yelp Review Polarity | 95.3 | 91.2 | 4.31% | 84.8 | 11.03% | 42.8 | 55.05% | 86.5 | 9.16% | 71.9 | 24.51% | 71.1 | 25.33% |
| Enron Spam Email | 95.6 | 85.5 | 10.56% | 69.0 | 27.84% | 76.4 | 20.13% | 85.1 | 11.03% | 78.7 | 17.68% | 75.4 | 21.14% |
| Mean | | | 10.97% | | 22.06% | | **48.58%** | | 11.05% | | **31.79%** | | **30.35%** |
| Median | | | 9.46% | | 25.44% | | **55.80%** | | 10.10% | | **30.19%** | | **29.11%** |
| Standard Deviation | | | 8.54% | | 11.53% | | 16.91% | | 7.10% | | 14.56% | | 12.93% |

Table 5: Effectiveness of WordBug on 8 Datasets using the Word-LSTM and Char-CNN model. Acc is the accuracy of the method and Decrease is the percent decrease of the accuracy by using the specified attacking method over the original accuracy. Word-LSTM uses Substitution transformer. All results are under maximum edit distance difference 30 ($\epsilon = 30$).

(30 out of 190) on average, and were able to get an average 73% decrease in accuracy using DeepWordBug.

To show that our method works on varied selection of maximum edit distance differences $\epsilon$, we present detailed experimental results for every datasets. The results are briefly summarized in Figure 6. In Figure 6, (a)(c) shows how the prediction accuracy decreases when relaxing the limitation on edit distance, using the result on the AG's News Dataset as an example. A full version of Char-CNN result are in Appendix Figure 14, and Word-LSTM result are in Appendix Figure 15. As expected, as the number of words attacked increases, accuracy decreases. However, we observe that even though the difference on edit distance is very small(e.g. $\epsilon = 5$), DeepWordBug generates many samples which successfully evade the classifier. Figure 6 (b)(d) summarize the average accuracy decrease on 8 datasets among all four DeepWordBug scoring functions and two baselines when $\epsilon = 30$, on Word-LSTM model and Char-CNN model respectively. On both models, DeepWordBug scoring functions largely outperforms baseline scoring functions.

An important result of our study is that regardless of the input type (word-level or character-level), our method is able to evade the classifiers. This is important for two reasons. First, this proves that Word-embedding based models are vulnerable to easy attacks by introducing unknown words. Second, character-level models are vulnerable to similar attacks, even though they are often used to improve generalization and not be affected by misspellings [30].

**Results from Baseline Comparisons:**
We can clearly see that our scoring method for finding important words is effective because randomly choosing words to change (i.e., "Random" in Table 5) has little influence on the final result.

Although the Gradient algorithm is not a black-box method and despite the fact that it has access to extra information, our scoring method still achieves better results. This difference is most likely attributable to the fact that the gradient is not a good measurement of token importance.

### 3.4 Transferability

An important property of adversarial image samples is their transferability. Adversarial samples that are generated for a certain model but can also successfully fool another DNN model on the same task are considered transferable.

We tested our attack for transferability on adversarial samples using the Combined Scoring function and the Substitution Transformer. In these experiments, we hold the maximum difference on edit distance ($\epsilon$) to a constant 30 for each sample. We evaluate the transferability using two different types of Word-LSTM models, one using a Uni-directional LSTM (LSTM), and one using a Bi-directional LSTM (BiLSTM). Additionally, we use two different types of word embeddings: LSTM1/BiLSTM1 are trained with randomly-initialized embedding, and LSTM2/BiLSTM2 are trained with GLoVE word embeddings [22].

Figure 7 shows the accuracy results from feeding adversarial sequences generated by one model to another model on the same task. The results show that the target model accuracy in these circumstances is reduced from around 90% to 20-50%. Most adversarial samples can be successfully transferred to other models, even to those models with different word embedding. This experiment demonstrates that our method can successfully find those words that are important for classification and that the transformation is effective across multiple models.

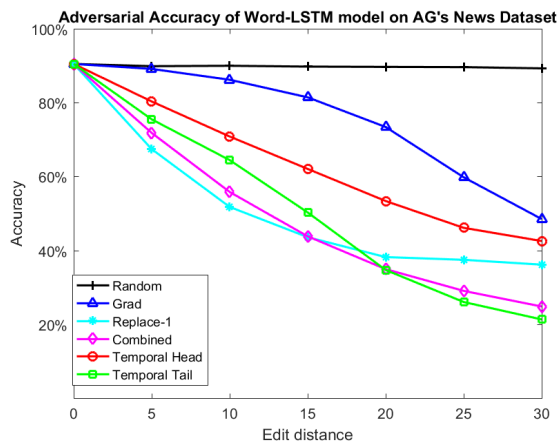
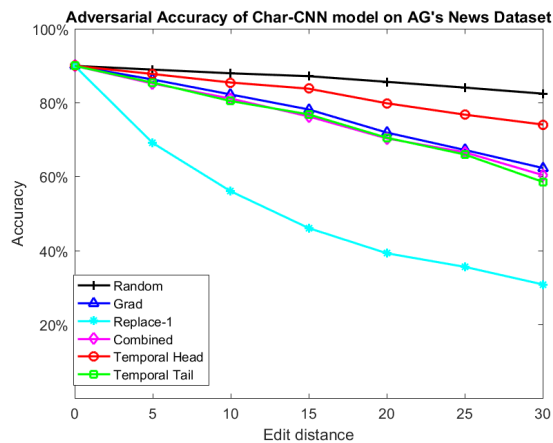

(a)

(c)

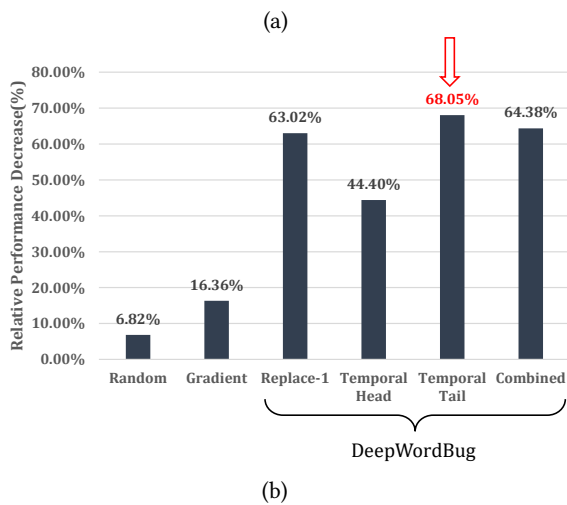
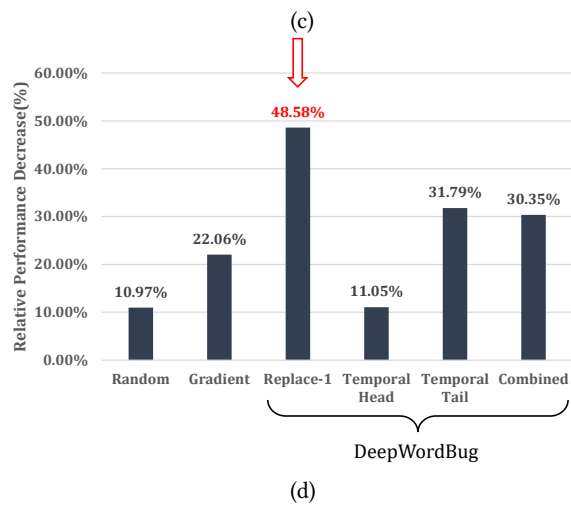

(b)

(d)

Figure 6: Experiment results of comparing baselines and DeepWordBug with different token scoring on two deep learning models. (a)(b) are the result of Word-LSTM, and (c)(d) are the result of Char-CNN. (a)(c) The variation of model accuracy on generated adversarial samples on AG's News Dataset. X axis represents the maximum allowed perturbation in edit distance (equal to number of words been modified in this case), and the Y axis corresponds to the test accuracy on adversarial samples generated using the respective attacking methods. (b)(d) A summary of DeepWordBug effectiveness: Average relative performance decrease (percentage of the accuracy when without the attacks) of the deep learning models when classifying adversarial samples generated from DeepWordBug and baselines. The higher a bar, the more effective. On eight large-scale text datasets, DeepWordBug (with temporal tail) leads to an average decrease of 68% from the original classification accuracy on the Word-LSTM models. On the Char-CNN models, DeepWordBug (with replace-1) leades to an average decrease of 48%. A full version of this figure can be found in Appendix (Fig 14 and Fig 15).

## 3.5 Comparing Transformer Functions

We also analyze the effectiveness of the four different transformer functions we propose. We use DeepWordBug with different transformer functions to attack a Word-LSTM model on the AG's News dataset. We use the Combined Score and the Substitution Transformer to generate adversarial samples, with the maximum edit distance difference of 30 ($\epsilon = 30$). The results are in Figure 8.

From this figure, we can conclude that varying the transformation function has a small influence on the attack performance of DeepWordBug, as all misspelled word are mapped to the same index "*unknown*" in the dictionary. Therefore, all the methods will produce the same sequence if the generated word is not in the vocabulary. As the probability of collision into another word is small, all the methods generate similar inputs. Swapping, however, when the edit distance difference is 2 instead of 1, has a worse performance in comparison to other methods under the same maximum difference of edit distance. The results show that varying the token scoring impacts the results more than changing the token transformer.

## 3.6 Influence of Dictionary Size

The dictionary size is a hyperparameter of word-based models. As our attack transforms words into *"unknown,"* the out of the vocabulary token, one may wonder whether the dictionary size has an influence on the effectiveness of DeepWordBug. In another experiment, we trained 4 Word-LSTM models with different dictionary sizes, ranging from 5,000 to 20,000. We used the Combined Score and the Substitution Transformer to generate adversarial samples, and the maximum edit distance difference is 30 ($\epsilon = 30$).

The results of applying our DeepWordBug attack on the four models of varying dictionary sizes are shown in Figure 9. The accuracy results between the different dictionary sizes we tested are very small. In other words, our method can work with different dictionary sizes.

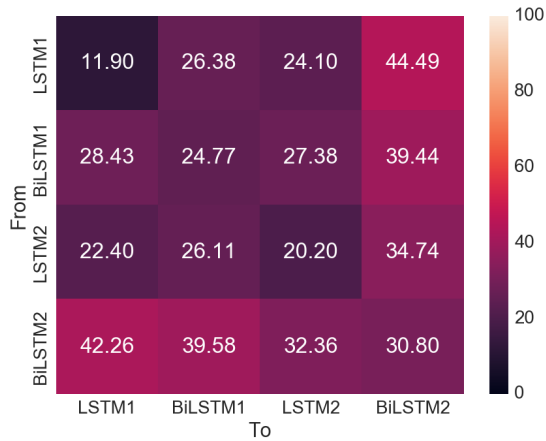

Figure 7: Heatmap that shows the transferability of DeepWordBug: The values are the accuracy of the target model tested on the adversarial samples.

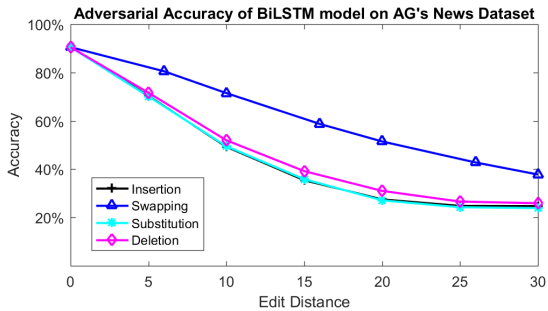

Figure 8: Comparing different token transformer strategies: Each curve represents the result of attacking the model with one transformer functions. X-Axis: Maximum edit distance difference Y-Axis: Model accuracy

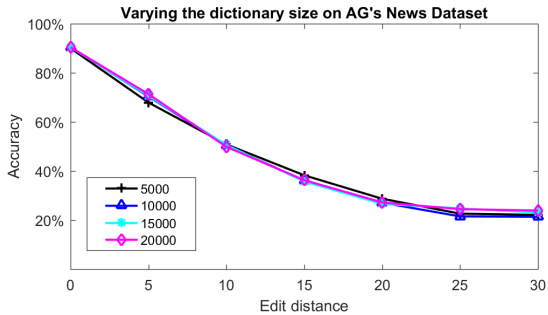

Figure 9: The accuracy of models when varying the size of the dictionary. X-Axis: Maximum edit distance difference. Y-Axis: Model accuracy.

## 4 CONNECTING TO PREVIOUS STUDIES

Research on attacking deep learning classifier starts from [6] and [13], when researchers show that malicious manipulation on input can cause machine learning based spam detectors to generate false positives. [14] propose the Good Word attack, which is a practical attack that adds positive, non-spam words into a spam message to evade machine learning classifiers. These attacks are designed for simple classifiers such as Naive Bayes, which work directly on different features. However, these methods do not have any guarantee on their performance on more complicated models such as deep learning classifiers. These methods also give no guarantee of the quality of the generated sample. For example, a spam message could become non-spam if too many "good words" were added to it. Such a message is not false positive, and thus not useful in the attack.

|  | Adversary | Distance | Space | Modifications |
|---|---|---|---|---|
| **Ours** | Black-box | Edit ($L_0$) | Input space | Swapping two characters |
| [21] | White-box | $L_\infty$ | Embedding space | Gradient + Projection |
| [25] | White-box | Num. words modified ($L_0$) | Input space | Complicated & Linguistic-driven |

Table 6: Summary of Relevant previous works.

In 2013, [28] proposes the concept of adversarial sample: imperceptible perturbation on images can fool deep learning classifiers. This discovery is interesting because small modifications guarantee the validity of generated samples. Compared to studies of adversarial examples on images, little attention has been paid on generating adversarial sequences on text. Papernot et al. applied gradient-based adversarial modifications directly to NLP inputs targeting RNN-based classifiers in [21]. The resulting samples are called "adversarial sequence," and we also adopt the name in this paper. The study proposed a white-box adversarial attack called projected Fast Gradient Sign Method and applied it repetitively to modify an input text until the generated sequence is misclassified. It first randomly picks a word, and then uses the gradient to generate a perturbation on the corresponding word vector. Then it maps the perturbed word vector into the nearest word based on Euclidean distance in the word embedding space. If the sequence is not yet misclassified, the algorithm will then randomly pick another position in the input.

Recently, [25] used the embedding gradient to determine important words. The technique used heuristic driven rules together with hand-crafted synonyms and typos. The method is a white-box attack since it accesses the gradient of the model. Another paper [12] measures the importance of each word to a certain class by using the word frequency from that class's training data. Then the method uses heuristic driven techniques to generate adversarial samples by adding, modifying, or removing important words. This method needs to access a large set of labeled data.

In summary, our approach differs from the previous approaches in that they are not applicable in a realistic black-box setting. Our method does not require knowing the structure, parameters or gradient of the target model, while previous methods do. Also, most previous approaches used heuristic-driven and complicated modification approaches. We summarize the differences between our method and previous methods on generating adversarial text samples in Table 6. Besides, [21] selects the words randomly (therefore, we have a "Random" baseline in our experiment); [25] selects words using gradient (therefore, we have a "Gradient" baseline in our experiment). Closely connected to [25], our method uses the edit distance at the sequence input space to search for the adversarial perturbations. Also, our token modification algorithm is simpler compared to [25].

## 5 DISCUSSIONS AND ANALYSES

### 5.1 Why WordBug Works

We present several examples of generated adversarial samples in Table 7. While the readability of our adversarial samples is subjective to the reader, we believe that they can be well understood by most readers, thus resulting in valid adversarial samples.

The key idea we utilized in DeepWordBug is that misspelled words are usually viewed as "*unknown*" by the deep-learning models. In contrast, humans in general can decipher the originally intended word. Psychologists have shown that people can accurately read paragraphs containing words constructed with swapped letters with relatively small time-cost of around 11% slowdown in reading speed [23]. Current deep-learning NLP models have not achieved the human-level of processing and understanding of natural language inputs. Though some deep-learning models use complicated feature extraction structure that can potentially catch the semantic expression of the words, they still fail to model the morphological similarity among words.

Following the algorithm, DeepWordBug can generate adversarial samples whose only difference to its seed sample are few letter-level modifications. Some may argue that the DeepWordbug adversarial samples are not truly adversarial samples, because the modifications are "perceptible" to human. However, DeepWordBug samples follows the definition given by Equation 1. More importantly, DeepWordBug samples can be understood by human to have the same meaning with its seed sample. Therefore, it should have the same prediction result to its seed sample. As many of samples are wrongly predicted into another class by the deep learning classifier, such samples are deliberately generated corner cases of the machine learning classifier, which are essentially adversarial samples.

### 5.2 Transferability

Recently, researchers found an adversary can transform black-box attacks to a white-box attack by utilizing transferability. [19] proposes an algorithm that queries the black-box target to gather information and build a local model. After the local model is built, [19] performs a white-box attack, by assuming the transferability of adversarial samples; that is, generated adversarial samples on the local model could also evade the target classifier.

While this local-model method can also be applied to generate black-box attacks on the text scenario, DeepWordBug is still a better way to generate adversarial samples in a local model. According to the results in Figure 15 and Figure 14, our method works better than the much-better-informed white-box baseline that uses gradient information. We believe that the token ranking techniques we explore on text adversarial sequences can shed light on other black-box techniques in the future.

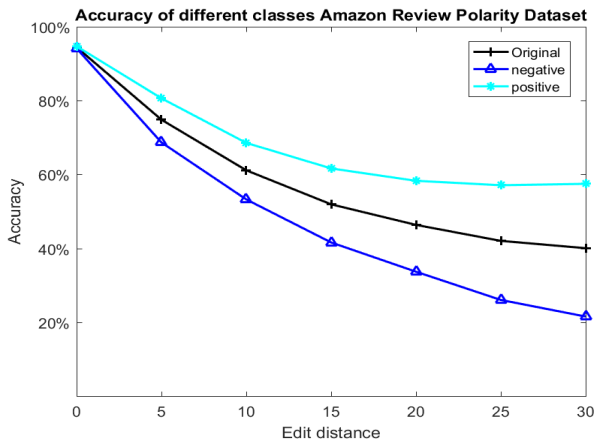

Figure 10: Performance of the attack on different classes of samples in the dataset. Each color represents one class of samples. X-Axis: Maximum edit distance difference. Y-Axis: Accuracy.

### 5.3 Performance of the attack on different classes of inputs

We also studied the performance of the attack on different classes of samples; that is, does our attack works better on certain classes than other classes? We present Figure 10 to answer this question. We attack the Word-LSTM model using the Combined Score in this experiment. From Figure 10, we conclude that our attack works differently on different classes: The accuracy on negative samples has been reduced to 20%, but the accuracy on the positive samples is still 60%.

### 5.4 Bias in the dataset

In the previous analysis, we assume "*unknown*" is not biased; that is, the word "unknown" is not favored by a typical class of inputs and means nothing to the machine learning model. However, in the practice, it is often the case that "*unknown*" is biased toward a certain class. In that case, by changing a certain word to be "*unknown*" the prediction will be biased towards that class, which contradicts our intended use of "*unknown*".

We studied the bias of "*unknown*" in the Enron Spam Dataset, as shown in Figure 11(a). From the figure, we can see that the "*unknown*" token is highly biased toward spam Emails, as 70% of "*unknown*" occurs in spam Emails. Therefore, our attack works better to make a non-spam Emailto be predicted as spam. This may answer the question of why our attack works differently among different sample classes.

We also studied the bias of tokens in the character-level models. Since our character transformer changes characters to the SPACE character, we searched for a bias in SPACE. Figure 11(b) shows the character level distribution. Comparing to "*unknown*," the SPACE symbol is less biased, as SPACE symbol occurs frequently in both spam Emails and non-spam Emails.

### 5.5 Rejecting Low Probability Classifications

When evaluating the accuracy of adversarial samples, we choose the class with the highest probability output. A logical question around the accuracy is how strongly the adversarial samples evade the classifier. In other words, does the model classify the wrong answer with strong confidence (give a wrong output with high probability) in the adversarial sample, or is the classification unsure (give a rather uniform output probability among the answers). In the second case, a possible defense is to treat low probability outputs as noise and reject the sample. However, we show that our adversarial samples will lead the model to give a high output probability on the wrong prediction.

Table 7: Examples of generated adversarial samples: The red part indicates the difference to the original message. In the character-level models, we represent the added space characters as "_".

| Data - Model | | Message | Prediction score | Prediction |
|---|---|---|---|---|
| Enron Spam - Word-LSTM | Original | Subject: breaking news. would you ref inance if you knew you ' d save thousands ? we ' ll get you lnterest as low as 3 . 23 % . | | |
| | Processed inputs | subject breaking news would you ref [OOV] if you knew you d save thousands we ll get you [OOV] as low as ... | 1.00 | spam |
| | DeepWordBug | subject breaking nwes would you ref [OOV] if you knew you d save thuosands we ll get you [OOV] as low as âĂę | 0.14 | non-spam |
| Enron Spam - Word-LSTM | Original | Subject: would you have an objection to mapping the east texas gas system to the carthage curve as opposed to the texoma curve ? | | |
| | Processed inputs | subject would you have an [OOV] to [OOV] the east texas gas system to the carthage curve as [OOV] to the texoma curve | 0.00 | non-spam |
| | DeepWordBug | sujbect woulg yuo hvae an [OOV] to [OOV] the east texma gsa ststem tp hte caethage curve bs [OOV] tn che texoam curve | 1.00 | spam |
| AG's News - Char-CNN | Original | saudi trial could alter pace of reform in a hearing room on the 11th floor of the high court of riyadh ... | 0.82 | World |
| | DeepWordBug | saudi trial co_ld alter pace of ref_rm in a hearing room on the 11th floor of the high court of riyadh ... | 0.81 | Sci/Tech |
| AG's News - Char-CNN | Original | sudan says u.n. sanctions would destroy society khartoum (reuters) - sudan said saturday that u.n. sanctions | 1.00 | World |
| | DeepWordBug | sudan says u_n. sanctions would destroy society khartoum (reuters) - sudan said saturday that u_n_ sanctions, | 0.66 | Sports |

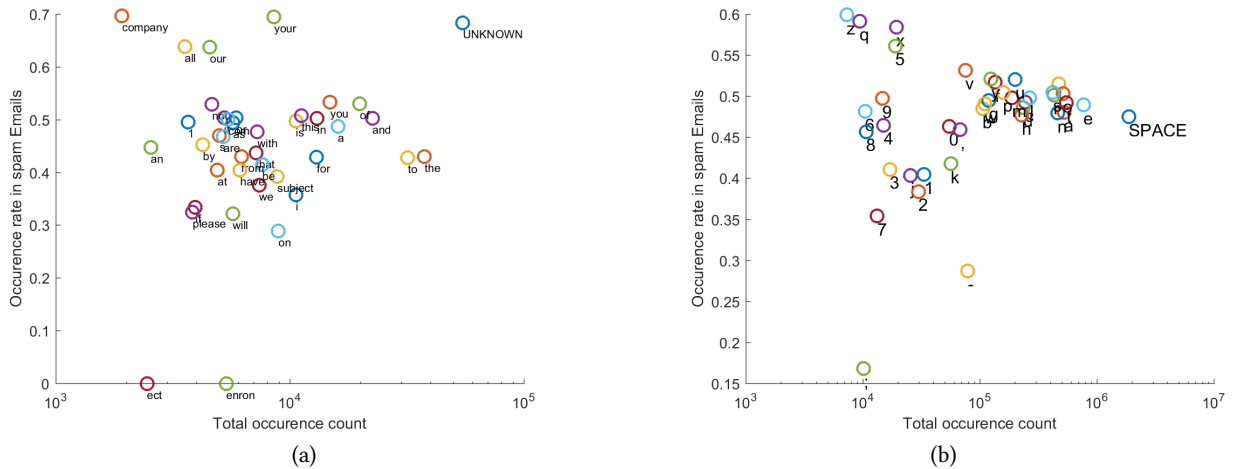

Figure 11: (a) Bias of "UNKNOWN" in Enron Spam Dataset. For both figures, X Axis is the total occurrence number of a certain word in the dataset, while y axis is the ratio of a certain word occurs in the spam Emails to the total occurrence. (b) Bias of "SPACE" character in Enron Spam Dataset when characters are used as the basic token.

We present two histograms to analyze the distribution of the output probability of the machine learning model on the fooled class in Figure 12. In the experiment, we used the Combined Score and the Substitution Transformer to generate adversarial samples, where maximum edit distance $\epsilon$ equals 30.

We show that 90% of the generated adversarial samples successfully make the deep learning model output a wrong answer with confidence probability larger than 0.9 in the Enron Spam Dataset. The result shows that generated adversarial samples are powerful in evading the classifier.

### 5.6 Adversarial training

Adversarial training is a technique to improve the deep learning model with adversarial samples[7]. In another experiment we show that our adversarial sample could help to improve the deep learning models to defend future attacks. We generated 20,000 adversarial samples on a subset of training data, and created a dataset with those adversarial samples and their corresponding original samples. We trained the model with the dataset for 10 epochs. While the accuracy of the model on normal samples slightly decreases, on adversarial samples the accuracy rapidly increases from around 12% (before the training) to 62% (after the training). The result is shown in Figure 13. The learning rate is set to be 0.0005 in this experiment.

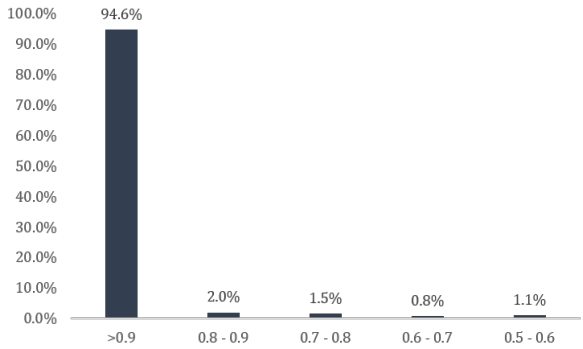

Figure 12: How strong the machine learning model will believe the wrong answer lead by the adversarial sample, the x-axis are the confidence range and the y-axis are the probability distribution. The result is generated using Word-LSTM model on the Enron Spam Dataset (Number of classes = 2), with edit distance maximum $\epsilon = 30$

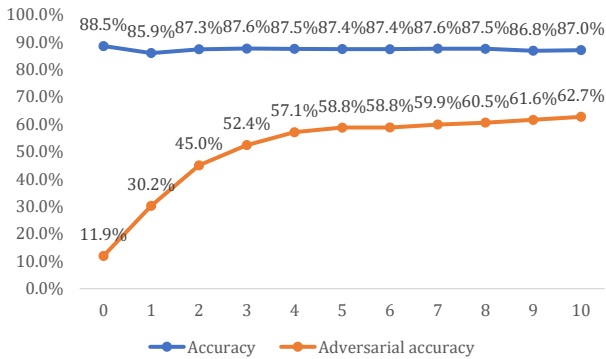

Figure 13: Adversarial training: The curve of the test accuracy on original samples and adversarial samples after epochs of adversarial training.

### 5.7 Autocorrection

A natural way to evaluate our proposed attacking method is to use an autocorrector for comparison. We tested the generated Deep-WordBug adversarial samples on a revised Word-LSTM model on which an autocorrector is applied before the input. The autocorrector we use is Python autocorrect 0.3.0 package. The result is shown in Table 8.

From the result in Table 8, we observe that while spellchecker reduces the performance of the adversarial samples, the reduction amount depends on which transformer function is chosen. Stronger attacks such as removing 2 characters in every selected word still can successfully reduce the accuracy of target deep learning model to 34%, which is close to a random guess in a 4-class prediction.

|  | Original | Attack ($\epsilon = 30$) | Autocorrector |
|---|---|---|---|
| Swapping-1 | 88.45% | 14.77% | 77.34% |
| Substitution-1 | 88.45% | 12.28% | 74.85% |
| Deletion-1 | 88.45% | 14.06% | 62.43% |
| Insertion-1 | 88.45% | 12.28% | 82.07% |
| Substitution-2 | 88.45% | 11.90% | 54.54% |
| Deletion-2 | 88.45% | 14.25% | 33.67% |

Table 8: Model accuracy with autocorrector.

There are several issues about using an autocorrector in practice. First, an autocorrector will add enormous overhead to the model. Autocorrection does not support batch processing, which takes a long time to process huge number of inputs. Second, the attackers can exploit the spell checker to make the situation even more dangerous. For example, attackers can hide malicious information in a deliberately created misspelling, and since such misspelling will be cleaned by the autocorrector, the machine learning model has no way to distinguish it from a normal message.

## 6 CONCLUSION

In this paper, we targeted a vulnerability with deep learning models for text classification. We present a novel framework, DeepWordBug, which can generate adversarial text sequences that can mislead deep learning models by exploiting this vulnerability. Our method has the following advantages:

- Black-box: DeepWordBug generates adversarial samples in a pure black-box manner.
- Performance: While constraining the edit distance difference of the adversarial sample, DeepWordBug achieves better performance compared to baseline methods on eight NLP datasets across two state-of-the-art deep learning architectures: Word-LSTM and Char-CNN.

Our experimental results indicate that DeepWordBug results in a 68% decrease on average from the original classification accuracy for a word-level LSTM model and 48% decrease on average for a character-level CNN model, both of which models are state-of-the-art.

We also demonstrated several important properties of DeepWordBug through experiments. First, the adversarial samples generated on one model can be successfully transferred to other models, reducing the target model accuracy from around 90% to 20-50%.

Second, our experiments show that the number of words in the dictionary used for inputs to the deep learning model does not affect our result.

Lastly, by using the samples generated by DeepWordBug in training data, the model accuracy on generated adversarial samples increases from 12% to around 62%.

# 7 APPENDIX

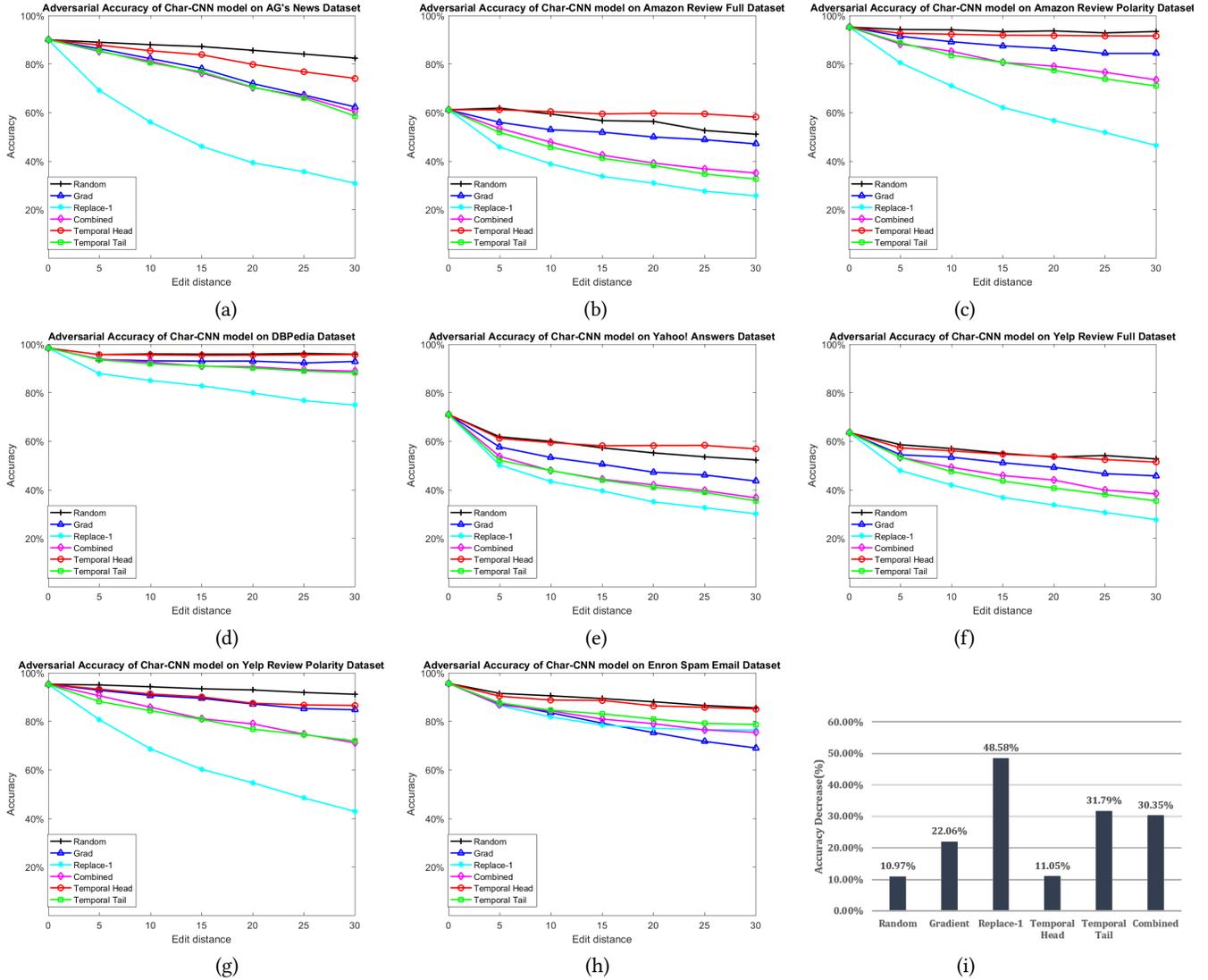

Figure 14: Experiment results of comparing baselines and DeepWordBug with different token scoring on Char-CNN model across eight datasets. (a) - (h) Model accuracy on generated adversarial samples, each figure represents the result on a certain dataset. The X axis represents the maximum allowed perturbation in edit distance (number of characters modified), and the Y axis corresponds to the test accuracy on adversarial samples generated using the respective attacking methods. (i) A summary of different attacking methods in the form of the percentage of average accuracy decrease. each bar represents one attacking method.

## 7.1 Influence of Different Transformers

As an extension to Figure 8, We conduct another experiment on all 8 datasets to study how varying different transformer functions affect the attack performance on a Word-LSTM model. We use combined score to generate adversarial samples, and maximum edit distance difference is $30(\epsilon = 30)$. The result is shown in Figure 16.

From these figures, we can conclude the same conclusion to 8 that varying transformation function have small influence on the attack performance. However, swapping costs twice edit distance distance, thus are worse than other methods.

## 7.2 Influence of dictionary size

In another experiment, we attack 4 Word-LSTM models trained with different dictionary sizes, ranging from 5,000 to 20,000, using DeepWordBug on all 8 datasets. We use combined score and the substitute transformer to generate adversarial samples, and maximum edit distance difference is $30(\epsilon = 30)$.

The results are shown in Figure 17. These result, together with 9, show that our method can work with different dictionary sizes.

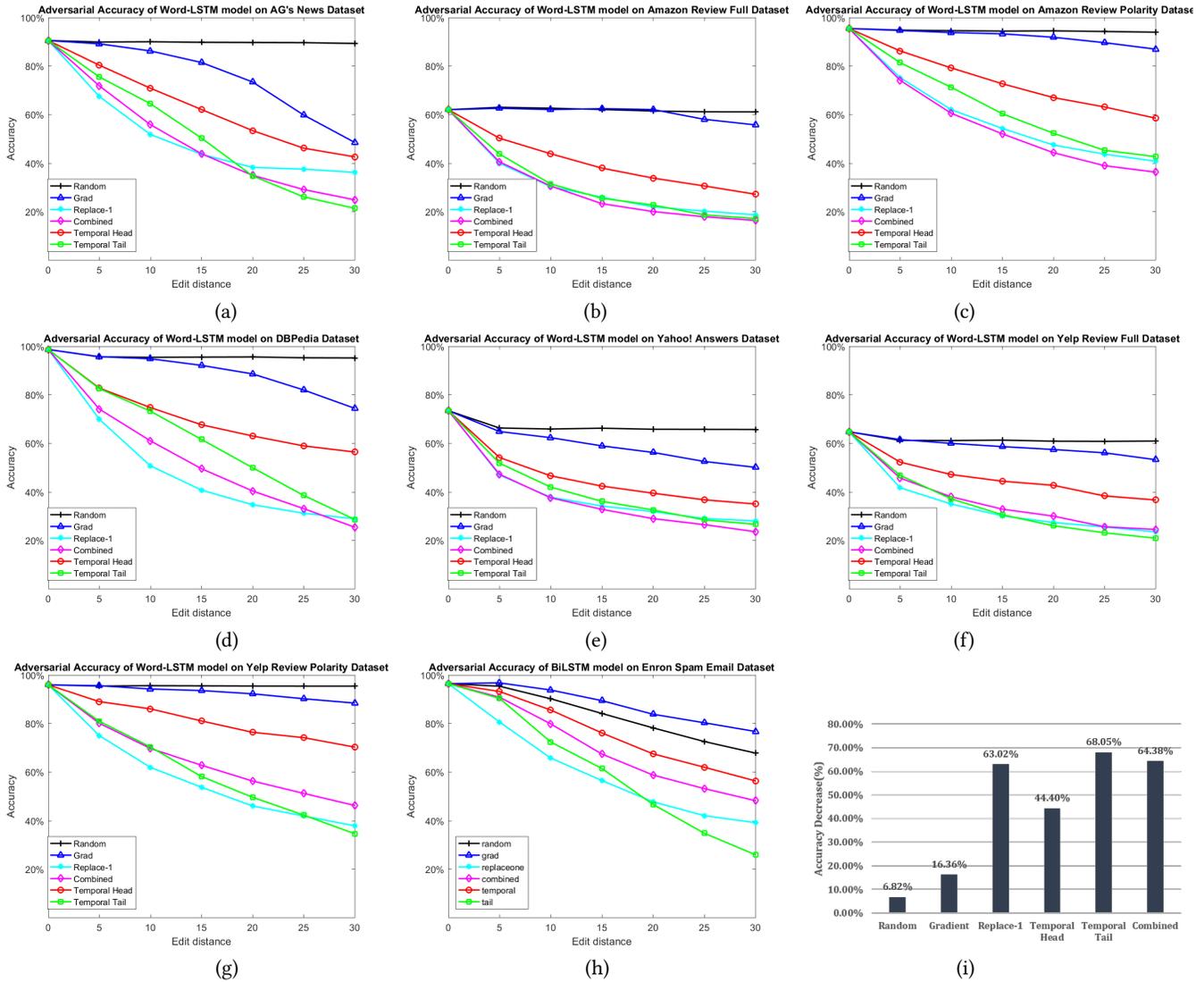

Figure 15: Experiment results of comparing baselines and DeepWordBug with different token scoring on word-LSTM model across eight datasets. (a) - (h) Model accuracy on generated adversarial samples, each figure represents the result on a certain dataset. The X axis represents the maximum allowed perturbation in edit distance (equal to number of words been modified in this case), and the Y axis corresponds to the test accuracy on adversarial samples generated using the respective attacking methods. (i) A summary of different attacking methods in the form of the percentage of average accuracy decrease. each bar represents one attacking method.

### 7.3 Classification Confidence on adversarial samples

We present histograms to analyze the distribution of the output probability of the machine learning model on the fooled class in Figure 18. In the experiment, we use the combined score and substitute transformer to generate adversarial samples, where the maximum edit distance difference $\epsilon = 30$.

This result is highly related to the number of the class in the prediction, which varies on different datasets. Together with the result in Figure 12, we can conclude that on every dataset, at least half of the adversarial samples have mislead the model to a probability 0.2 larger than a random guess (dashed line in the figures), which means generated adversarial samples are powerful to mislead models to believe a wrong answer with high confidence.

### 7.4 Bias in different datasets

Figure 11 shows the bias on the dataset, including both word models and character models.

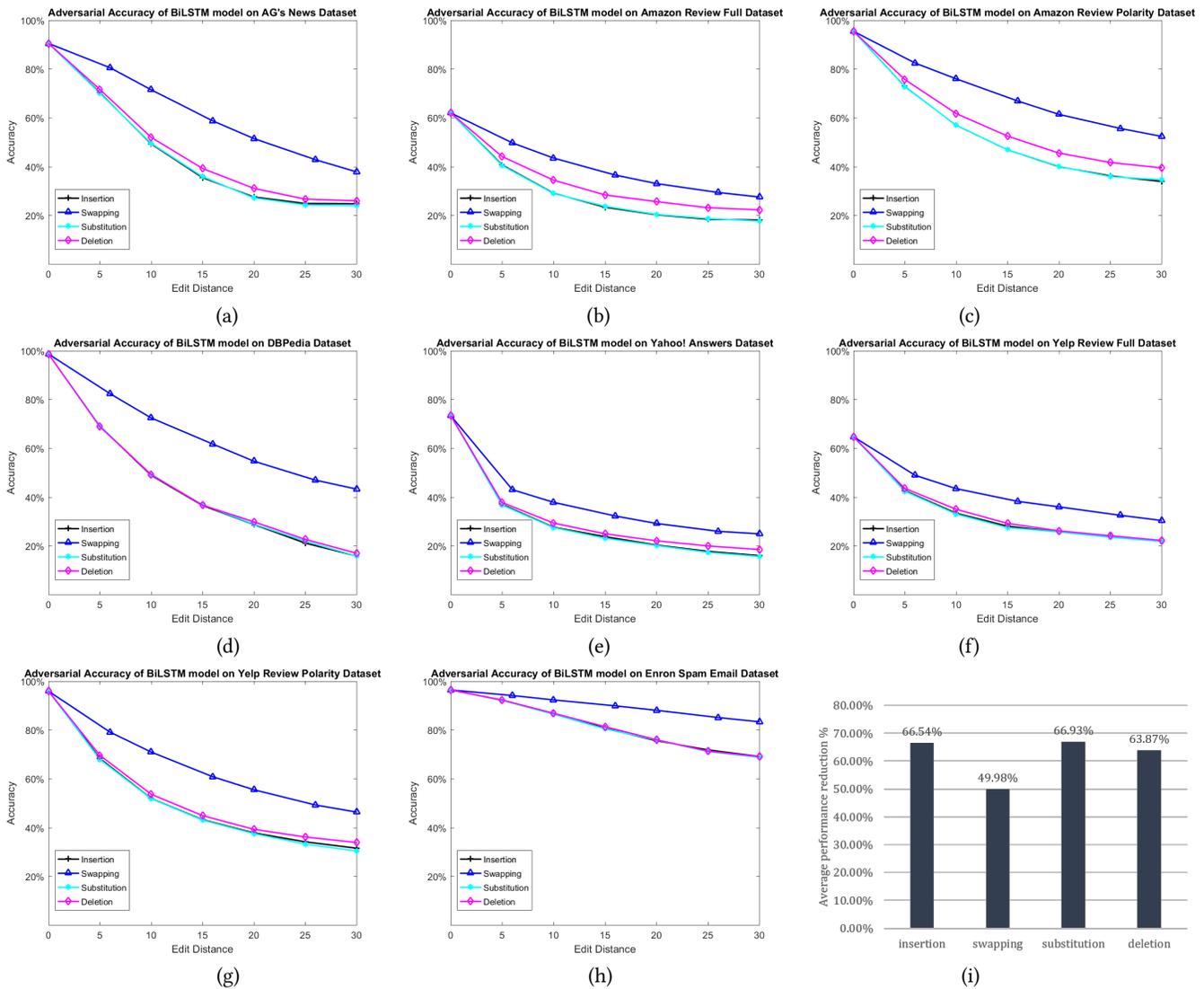

Figure 16: Comparing different transformer algorithms: From (a) to (h), each curve represents the result of attacking the model with one transformer functions. (i) A summary of different transformers in the form of the percentage of average accuracy decrease. X-Axis: Maximum edit distance difference. Y-Axis: Model accuracy.

Now we present the word occurrence ratio of frequent words, including "*unknown*", on all 8 dataset in Figure 19. From the result, we can conclude that large bias of unknown exists in some datasets (i.e., Enron Spam Dataset) while the bias is very small in other datasets (i.e., AG's News).

We also present the occurrence count of frequent letters, including the space symbol, on all 8 datasets in Figure 20. Comparing to the word level, our result shows that on the character level the data is less biased.

## 7.5 Attacking inputs of different classes

In another experiment, we study the performance of DeepWordBug attack on different classes of samples. In the experiment, we attack the Word-LSTM model use combined score to generate adversarial samples in this experiment.

We present the figures in the Figure 21. From the figure, we can see that our attack works differently on different classes. This phenomenon can be caused by the bias existed in the dataset, which we present in Figure 19.

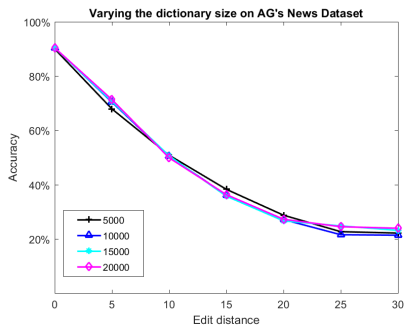 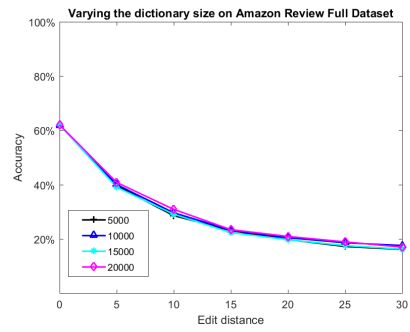 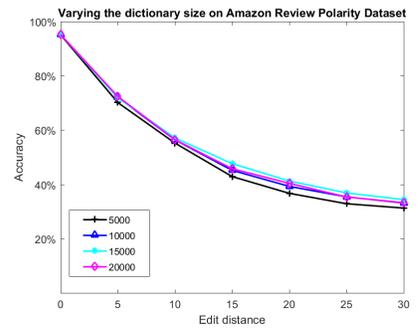
(a) (b) (c)

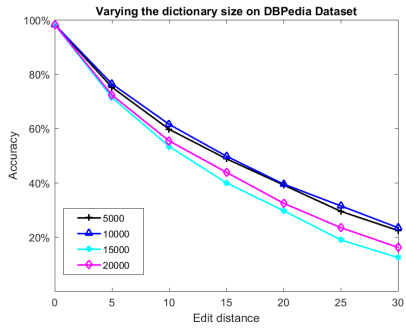 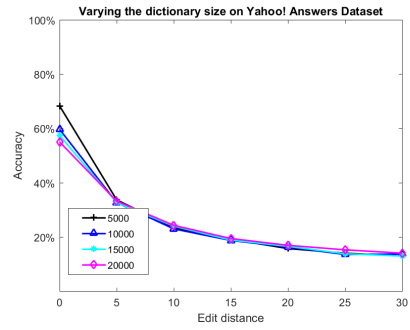 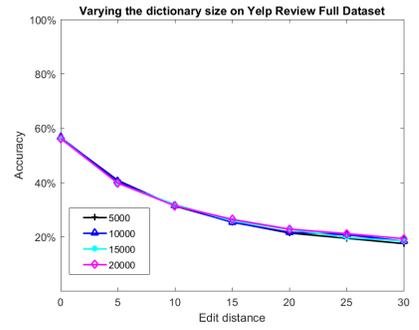
(d) (e) (f)

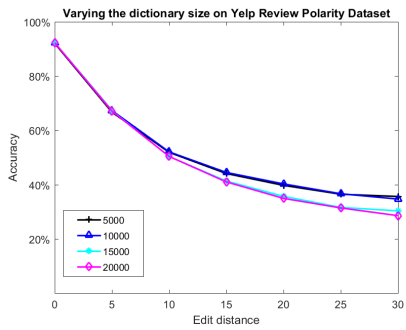 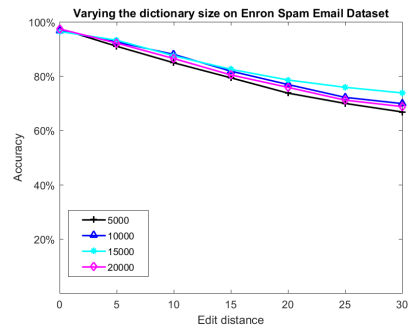 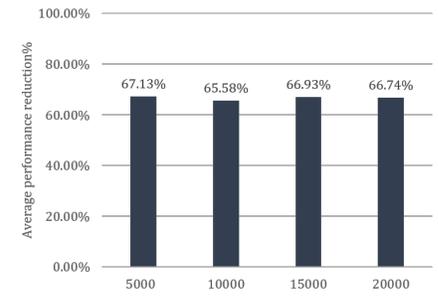
(g) (h) (i)

Figure 17: Effect of DeepWordBug on models with different dictionary sizes: From (a) to (h), each curve shows how our attacks work on models with different dictionary size on one dataset. (i) A summary of the attack on models with different dictionary size in the form of the percentage of average accuracy decrease. X-Axis: Maximum edit distance difference. Y-Axis: Model accuracy

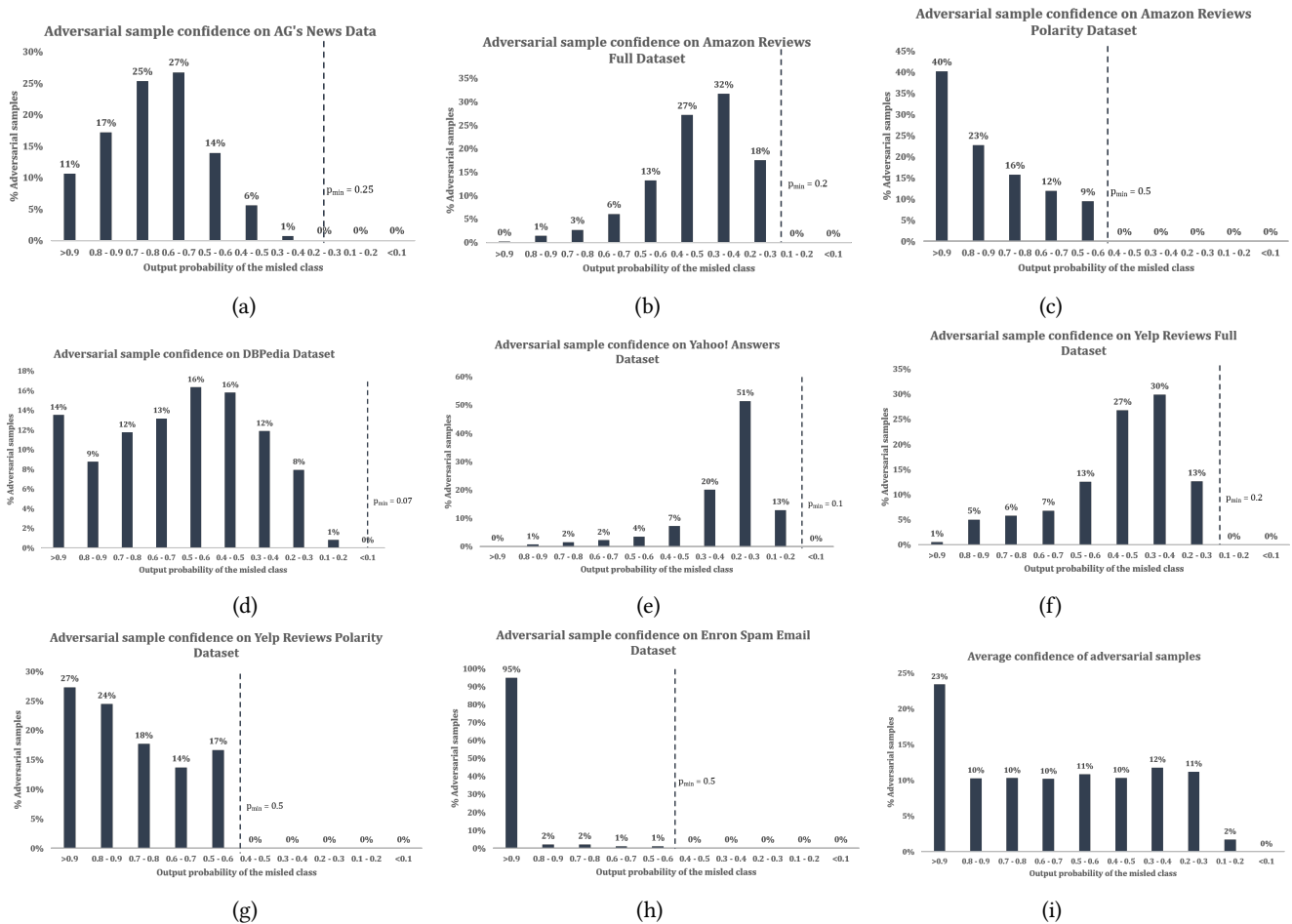

Figure 18: How strong the machine learning model will believe the wrong answer lead by the adversarial sample, the x-axis are the confidence range and the y-axis are the probability distribution. The dashed line shows for the base probability, as the probability should be larger than average. (a) - (h) The confidence probability distribution on different datasets (i) A summary of different transformers in the form of the percentage of average accuracy decrease.

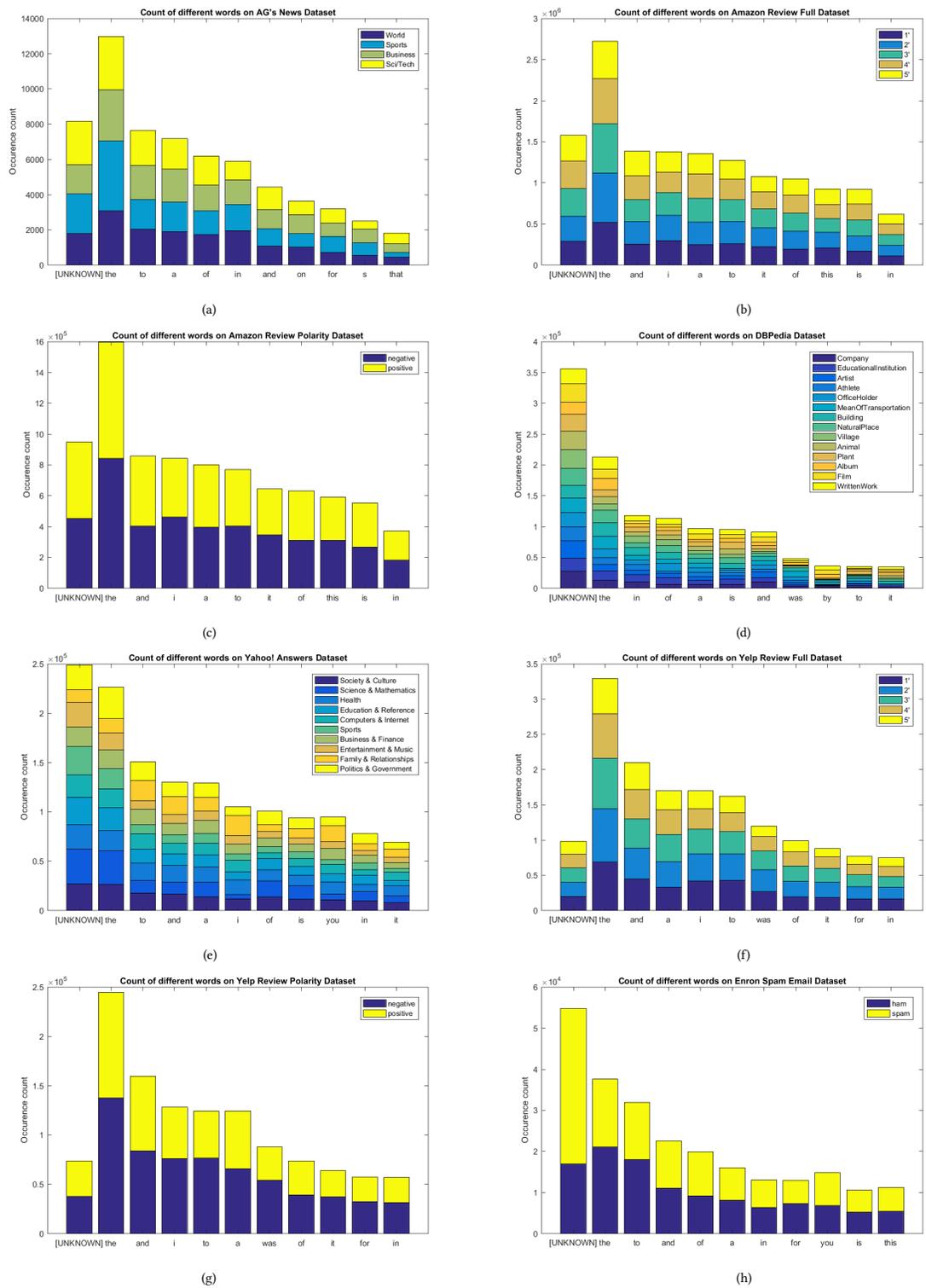

Figure 19: Word frequency ratio of frequent words on different classes of the dataset. Each color represents a different class in the data. From (a) to (h), each bar plot shows the how often a certain word occurs in one class on one dataset. Y-Axis: Word occurrence ratio, the result of every class sums up to 1

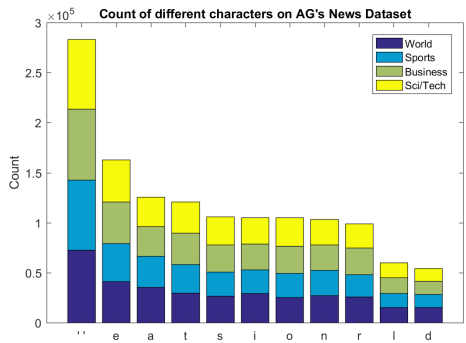

(a)

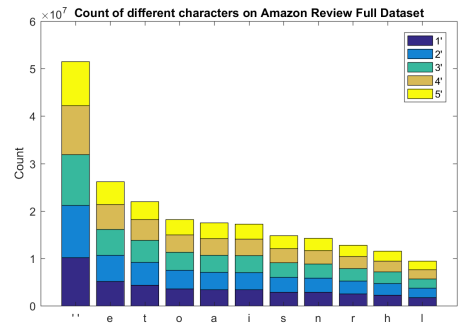

(b)

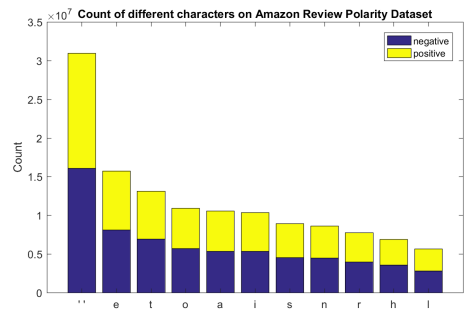

(c)

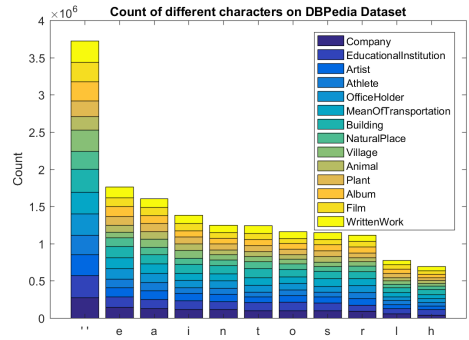

(d)

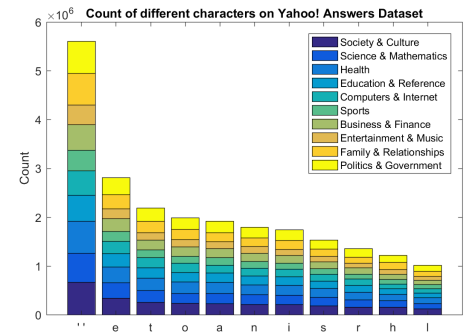

(e)

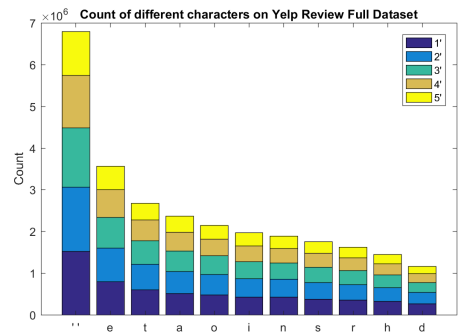

(f)

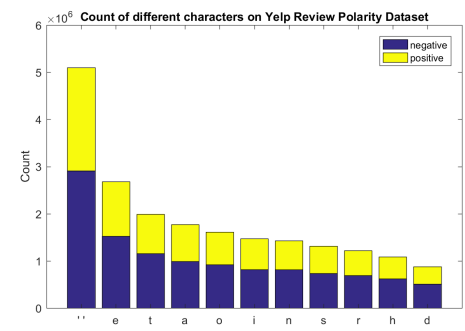

(g)

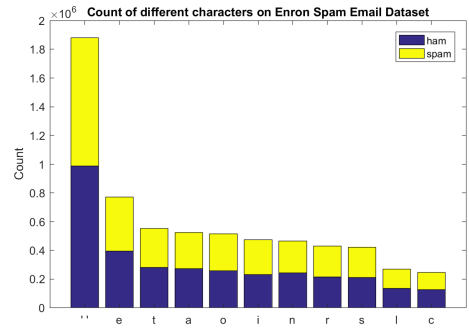

(h)

**Figure 20: Count of frequent characters on different classes of the dataset. Each color represents a different class in the data. From (a) to (h), each bar plot shows the count of frequent characters on one dataset. Y-Axis: Number of occurrence**

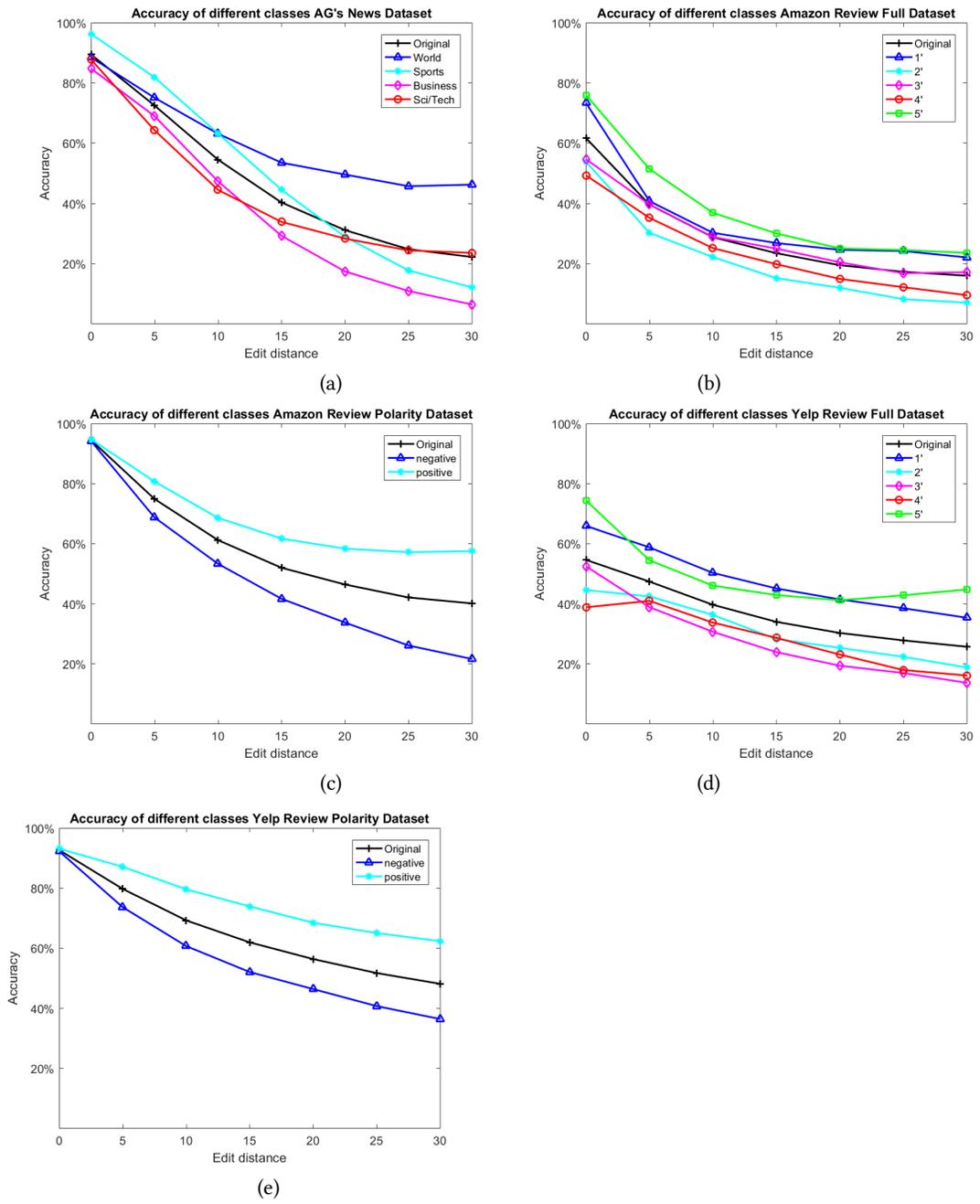

Figure 21: DeepWordBug attack on different classes of samples in the dataset. Each color represents the result of attack on one class of samples in the data. From (a) to (e), each plot shows the result of one dataset. X-Axis: Maximum edit distance difference. Y-Axis: Accuracy

21